%% file: main.tex
\pdfoutput=1

\documentclass[11pt]{article}

\usepackage[preprint]{ACL}
\usepackage{times}
\usepackage{latexsym}
\usepackage[T1]{fontenc}
\usepackage[utf8]{inputenc}
\usepackage{microtype}
\usepackage{inconsolata}
\usepackage{xcolor}         
\usepackage{graphicx}
\usepackage{booktabs}
\usepackage{bm}
\usepackage{IEEEtrantools}
\usepackage{color, colortbl}
\usepackage{xspace}
\usepackage{multirow,makecell}
\usepackage{amsmath}
\usepackage{amsfonts}       

\definecolor{loss-orange}{HTML}{FF9D45}
\definecolor{Highlight}{rgb}{0.89,0.89,0.94}
\newcommand{\chl}{\cellcolor{Highlight}}

\newcommand{\mymodel}{\textsc{EvoLlama}\xspace}
\newcommand{\mympnnmodel}{\textsc{EvoLlama} (ProteinMPNN+ESM-2)\xspace}
\newcommand{\mygearmodel}{\textsc{EvoLlama} (GearNet+ESM-2)\xspace}

\title{\mymodel: Enhancing LLMs' Understanding of Proteins via Multimodal Structure and Sequence Representations}

\author{
    Nuowei Liu\textsuperscript{1}\thanks{\ Equal Contribution},
    Changzhi Sun\textsuperscript{2}\footnotemark[1],
    Tao Ji\textsuperscript{3}, \\
    \textbf{Junfeng Tian\textsuperscript{4},}
    \textbf{Jianxin Tang\textsuperscript{5},}
    \textbf{Yuanbin Wu\textsuperscript{1}\thanks{\ Corresponding authors},}
    \textbf{Man Lan\textsuperscript{1}\footnotemark[2]} \\
    \textsuperscript{1}School of Computer Science and Technology, East China Normal University \\
    \textsuperscript{2}Institute of Artificial Intelligence (TeleAI), China Telecom \\
    \textsuperscript{3}School of Computer Science, Fudan University \\
    \textsuperscript{4}Xiaohongshu Inc \\
    \textsuperscript{5}Innovation Center for AI and Drug Discovery, School of Pharmacy, East China Normal University \\
    \texttt{nwliu@stu.ecnu.edu.cn}, \texttt{czsun.cs@gmail.com}, \texttt{taoji@fudan.edu.cn}, \\ \texttt{tianjunfeng@xiaohongshu.com}, \texttt{52274300028@stu.ecnu.edu.cn}, \\ \texttt{\{ybwu, mlan\}@cs.ecnu.edu.cn}
}

\begin{document}
\maketitle
\begin{abstract}
\input{docs/000abstract}
\end{abstract}

\section{Introduction}
\label{sec:intro}
\input{docs/010intro}

\section{Related Work}
\label{sec:related}
\input{docs/040related}

\section{Approach}
\label{sec:approach}
\input{docs/020approach}

\section{Protein Instruction-Following Data}
\label{sec:instruct}
\input{docs/015instruct}

\section{Experiments}
\label{sec:exp}
\input{docs/030experiment}

\section{Conclusion}
\label{sec:conclusion}
\input{docs/050conclusion}

\section*{Limitations}
\label{sec:limitations}
\input{docs/055limitations}

\bibliography{reference}

\appendix
\label{sec:appendix}
\input{docs/060appendix}

\end{document}

%% file: docs/000abstract.tex
Current Large Language Models (LLMs) for understanding proteins primarily treats amino acid sequences as a text modality.
Meanwhile, Protein Language Models (PLMs), such as ESM-2, have learned massive sequential evolutionary knowledge from the universe of natural protein sequences. 
Furthermore, structure-based encoders like ProteinMPNN learn the structural information of proteins through Graph Neural Networks.
However, whether the incorporation of protein encoders can enhance the protein understanding of LLMs has not been explored.
To bridge this gap, we propose \mymodel, a multimodal framework that connects a structure-based encoder, a sequence-based protein encoder and an LLM for protein understanding.
\mymodel consists of a ProteinMPNN structure encoder, an ESM-2 protein sequence encoder, a multimodal projector to align protein and text representations and a Llama-3 text decoder.
To train \mymodel, we fine-tune it on protein-oriented instructions and protein property prediction datasets verbalized via natural language instruction templates.
Our experiments show that \mymodel's protein understanding capabilities have been significantly enhanced, outperforming other fine-tuned protein-oriented LLMs in zero-shot settings by an average of 1\%-8\% and surpassing the state-of-the-art baseline with supervised fine-tuning by an average of 6\%.
On protein property prediction datasets, our approach achieves promising results that are competitive with state-of-the-art task-specific baselines.
We will release our code in a future version.

%% file: docs/010intro.tex
The rapid advancements in Natural Language Processing (NLP) have led to the development of Large Language Models (LLMs) that are capable of understanding and generating human language. These models such as GPT-3.5 \citep{openai2022chatgpt}, GPT-4 \citep{achiam2023gpt} and Llama \citep{touvron2023llama,touvron2023llama2,dubey2024llama3}, inherently possess a certain level of world knowledge and have demonstrated remarkable proficiency across a wide range of tasks. Recently, the field of Bioinformatics has seen the emergence of Transformer-based \citep{vaswani2017attention} Protein Language Models (PLMs) like ProtBert \citep{elnaggar2021prottrans} and ESM \citep{rives2021biological,lin2022language}. These sequence-based encoders are pre-trained on a large number of amino acid sequences to capture the functional information embedded within proteins. Moreover, structure-based encoders like ProteinMPNN \citep{dauparas2022robust} and GearNet \citep{zhang2022protein} utilize Graph Neural Networks to learn the structural information of proteins.

Despite the success of protein encoders and LLMs in their respective domains, a significant gap remains in integrating the knowledge from protein encoders into LLMs to address biological problems by leveraging the parametric knowledge of LLMs. Current LLMs treat amino acid sequences as a text modality \citep{pei2023biot5,fang2023mol}, potentially failing to leverage the rich structural and sequential information of proteins that protein encoders are designed to capture. Moreover, protein encoders face challenges in multi-task learning, making them unable to follow human instructions. Besides, the gap between protein encoders and LLMs leads to significant challenges in aligning different modalities, even between the primary and tertiary structures of proteins \citep{zhang2023systematic}.

To address the aforementioned challenges, we introduce \mymodel, a multimodal framework designed to integrate the capabilities of protein encoders with an LLM. \mymodel combines the ESM-2 \citep{lin2022language} protein sequence encoder, which captures sequential evolutionary knowledge from amino acid sequences, the ProteinMPNN \citep{dauparas2022robust} structure encoder that learns geometric features from 3D protein structures, a multimodal projector that aligns protein and text representations, and a Llama-3 \citep{dubey2024llama3} text decoder for generating natural language outputs.

We propose a two-stage training approach, and the experimental results demonstrate that \mymodel with zero-shot outperforms the baselines fine-tuned on the Mol-Instructions \citep{fang2023mol} dataset by an average of 1\%-8\% and surpasses the current state-of-the-art model with supervised fine-tuning by an average of 6\%.
Additionally, on protein property prediction tasks based on the PEER benchmark \citep{xu2022peer}, \mymodel shows promising results that are competitive with task-specific baselines. 

Our contributions are listed as follows:

\begin{itemize}
    \item \textbf{Leverage multimodal representations of protein structures and sequences.}
    We align protein structure and sequence representations with LLM text modalities, bridging the gap in limitations of protein encoders that are unable to directly exploit the advanced capabilities of LLMs. Our approach enhances LLMs’ understanding of proteins, leveraging their parametric knowledge to address biological problems and laying a foundation for future research on incorporating a broader range of biomolecular modalities.
    \item \textbf{Multi-task learning and instruction following capability.}
    We implement a two-stage training approach. After projection tuning, \mymodel can follow various human instructions and solve downstream tasks in zero-shot settings. During the supervised fine-tuning stage, experiments demonstrate that tasks in the PEER benchmark have few interrelations and do not negatively affect each other when multi-task fine-tuning is employed.
    \item \textbf{Plug-and-play architecture and efficient fine-tuning approach.}
    Different protein encoders and LLMs can be used in our plug-and-play architecture. Extensive experiments demonstrate that the projection tuning stage can be optional, with the frozen LLM parameters during supervised fine-tuning significantly reducing trainable parameters. Additionally, we introduce a simple yet effective fusion method to align structure and sequence representations, improving efficiency during both training and inference.
\end{itemize}

%% file: docs/040related.tex
\paragraph{Protein-oriented LLMs}

BioT5 \citep{pei2023biot5} and BioT5+ \citep{pei2024biot5+} captures the underlying relations and properties of bio-entities such as molecules and proteins.
ProLLaMA \citep{lv2024prollama} introduces a training framework to transform a LLM into a multi-task protein LLM, focusing on protein sequence generation and superfamily prediction tasks.
InstructProtein \citep{wang2024instructprotein} utilizes a knowledge graph based-generation framework to construct instructions.
These methods utilize text-format protein sequences while \mymodel focuses on leveraging multimodal representations of proteins.
Prot2Text \citep{abdine2024prot2text} directly fuses the structure and sequence representations as inputs into the multi-head cross-attention module within the Transformer decoder. Compared to Prot2Text, \mymodel maps the structure and sequence features into language embedding tokens, enabling it to handle PPI prediction tasks, which typically requires two proteins as inputs.
Additionally, Prot2Text is designed to generate protein descriptions rather than handle various protein-oriented tasks, whereas \mymodel can follow human instructions, even in zero-shot settings.
Further related work is discussed in Appendix~\ref{sec:related-more}.

\paragraph{Protein Representations}
BERT-based PLMs such as ProtBert \citep{elnaggar2021prottrans}, ProteinBERT \citep{brandes2022proteinbert} and ESM \citep{rives2021biological,lin2022language} learn protein sequence representations through Masked Language Modeling objective.
\citet{gligorijevic2021structure} propose a Graph Convolutional Network to encode protein structures.
\citet{zhang2022protein} present a protein graph encoder to learn protein geometric features.
Apart from these sequence-based protein encoders, some work employ Graph Neural Networks to learn the geometric features of proteins.
\citet{dauparas2022robust} introduce a protein sequence design method based on Message Passing Neural Network with 3 encoder and 3 decoder layers. We adopt the encoder layers of ProteinMPNN as a structure-based encoder in our approach.
GearNet \citep{zhang2022protein} performs relational message passing on protein residue graphs for protein representation learning.
While these methods effectively learn the protein representations through sequences or structures, they do not utilize natural language with knowledge of protein properties. Therefore, \mymodel aligns protein and text representations to enhance LLM's understanding of proteins.

%% file: docs/020approach.tex
\begin{figure*}[!htb]
  \centering
  \includegraphics[width=\textwidth]{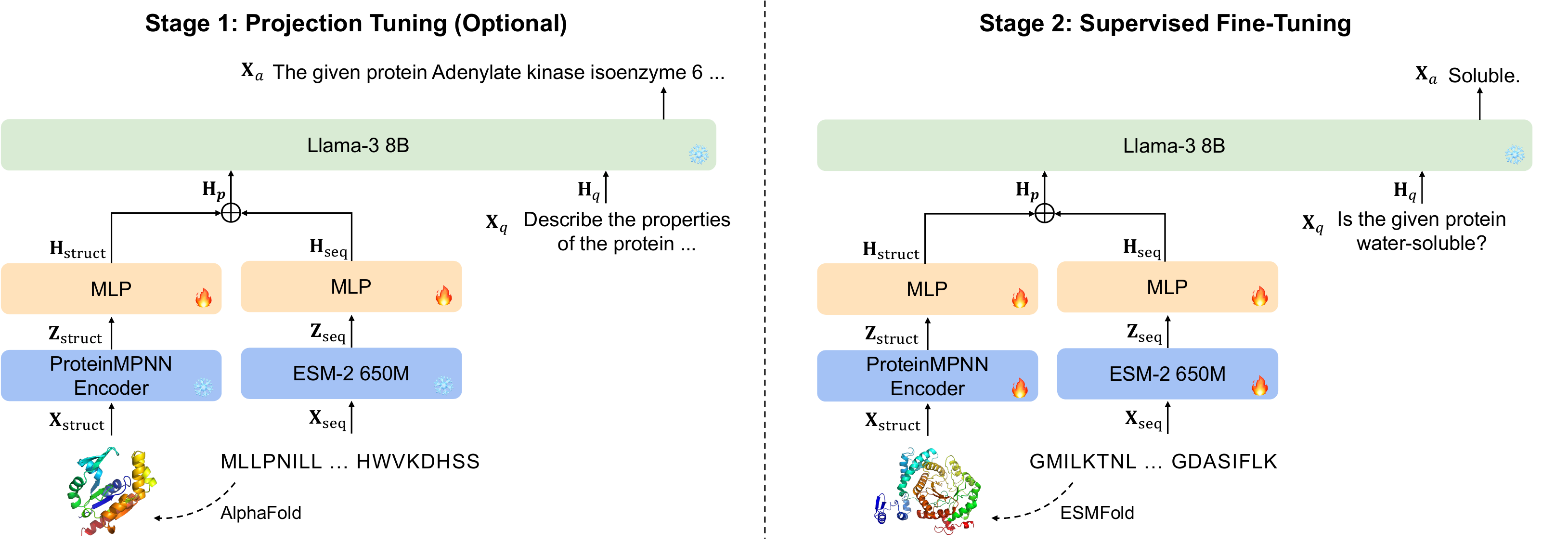}
  \caption{Overall architecture and the training pipeline of the \mymodel.} 
  \label{fig:architecture}
\end{figure*}

\mymodel aims to align protein information from both pre-trained structure-based and sequence-based protein encoders with an advanced LLM. 
Both language and protein models are open-sourced.
We target to bridge the gap between the protein encoders and LLM using MLP projection layers (Sec.~\ref{sec:architecture}), with an overview of our model displayed in Fig.~\ref{fig:architecture}.
To achieve an effective \mymodel, we propose a two-stage training approach (Sec.~\ref{sec:training}).
The initial stage involves pre-training the model on a large collection of aligned protein-text pairs to acquire protein language knowledge. 
In the second stage, we fine-tune the model with the high-quality protein-text dataset to enhance generation reliability and usability.

\subsection{Architecture}
\label{sec:architecture}

In this section, we will introduce the overall \mymodel in three parts: the protein encoders, the projection layer and the language decoder.

\paragraph{Protein Encoders}
Given the input amino acid sequence $\mathbf{X}_\text{seq}$, we consider the pre-trained protein encoder ESM-2 ~\citep{lin2022language}, which provides the protein feature $\mathbf{Z}_\text{seq} = \mathtt{SeqEncoder}(\mathbf{X}_\text{seq})$. The 3D structure of the given amino acid sequence is predicted using AlphaFold-2 \citep{jumper2021highly} or ESMFold \citep{lin2022language}. A protein structure encoder, such as ProteinMPNN and GearNet, is used to extract the feature \(\mathbf{Z}_{\text{struct}} = \mathtt{StructEncoder}(\mathbf{X}_{\text{struct}})\).

\paragraph{Projection Layer}
To map the outputs of the protein encoders into the same space as the text features from word embedding,
we apply an MLP to convert $\mathbf{Z}_{\text{seq}}$ and \(\mathbf{Z}_{\text{struct}}\) into language embedding tokens $\mathbf{H}_{\text{seq}}$ and \(\mathbf{H}_{\text{struct}}\) separately, which have the same dimensionality of the word embedding space in the language model:
\begin{IEEEeqnarray*}{c}
    \begin{cases}
      \mathbf{H}_{\text{seq}} = \mathtt{MLP}_{\text{seq}}(\mathbf{Z}_\text{seq}), \\ \mathbf{Z_\text{seq}} = \mathtt{SeqEncoder}(\mathbf{X}_\text{seq}), \\
      \mathbf{H}_{\text{struct}} = \mathtt{MLP}_{\text{struct}}(\mathbf{Z}_\text{struct}), \\ \mathbf{Z_\text{struct}} = \mathtt{StructEncoder}(\mathbf{X}_\text{struct})
    \end{cases}
\end{IEEEeqnarray*}
Furthermore, since both structure-based and sequence-based protein encoders extract features based on residue positions, the lengths of their feature representations are dependent solely on the length of the amino acid sequence. Therefore, we fuse the structure and sequence features by employing an element-wise addition of the corresponding residue features. The fused protein representations \(\mathbf{H}_p = \mathbf{H}_{\text{seq}} \oplus \mathbf{H}_{\text{struct}}\) reduces the protein embedding tokens by half, significantly decreasing the training and inference latency.
Note that our simple projection scheme is lightweight and cost-effective, which allows us to iterate data centric experiments quickly.
We leave exploring possibly more effective and sophisticated architecture designs for \mymodel as future work.

\paragraph{Language Decoder}
Given the protein structure $\mathbf{X}_{\text{struct}}$, amino acid sequence \(\mathbf{X}_{\text{seq}}\) and the fused projected embeddings $\mathbf{H}_p$,
we have conversation data $(\mathbf{X}_q, \mathbf{X}_a)$, where \(\mathbf{X}_q\) and \(\mathbf{X}_a\) represent the protein-related question and its corresponding answer, respectively.
We organize them as a sequence and perform instruction-tuning of the LLM on the prediction tokens, using its original auto-regressive training objective.
Specifically, we compute the probability of generating target answers $\mathbf{X}_a$ by:
\begin{IEEEeqnarray}{c}
\label{eq:llm}
p(\mathbf{X}_a | \mathbf{X}_{\text{struct}}, \mathbf{X}_{\text{seq}}, \mathbf{X}_q) \nonumber \\ 
= \prod_{i = 1}^{|\mathbf{X}_a|}
p_\theta(\mathbf{X}_{a, i}| \mathbf{X}_{\text{struct}}, \mathbf{X}_{\text{seq}}, \mathbf{X}_q, \mathbf{X}_{a, <i})
\end{IEEEeqnarray}
where $\theta$ is the trainable parameters.
\mymodel model design is compatible with any off-the-shelf GPT-style pre-trained LLM. 
\mymodel adopts Llama-3 8B \citep{dubey2024llama3} for further training. 
A causal mask is applied to all the attention operations, including the attention between protein features $\mathbf{H}_p$.

\subsection{Training}
\label{sec:training}
As illustrated in Fig.~\ref{fig:architecture}, the training process of the \mymodel model consists of two stages: projection tuning and supervised fine-tuning, with the first stage being optional.

\paragraph{Stage 1: Projection Tuning}
We keep both the protein encoders and LLM weights frozen, and maximize the likelihood of Eq.~\ref{eq:llm} with the parameters of projection MLP only (Fig. \ref{fig:architecture}(a)). 
In this way, the protein features $\mathbf{H}_p$ can be aligned with the pre-trained LLM word embedding. 
This stage can be understood as training a compatible protein projector for the frozen LLM.

\paragraph{Stage 2: Supervised Fine-Tuning}
To efficiently fine-tune \mymodel and preserve the internal knowledge of the LLM, its parameters are frozen during this stage. Continuing to update the pre-trained weights of the projection layers and protein encoders helps \mymodel learn more protein knowledge and enhance its instruction following capability.

%% file: docs/015instruct.tex
\begin{figure}[!htb]
    \centering
    \includegraphics[width=\linewidth]{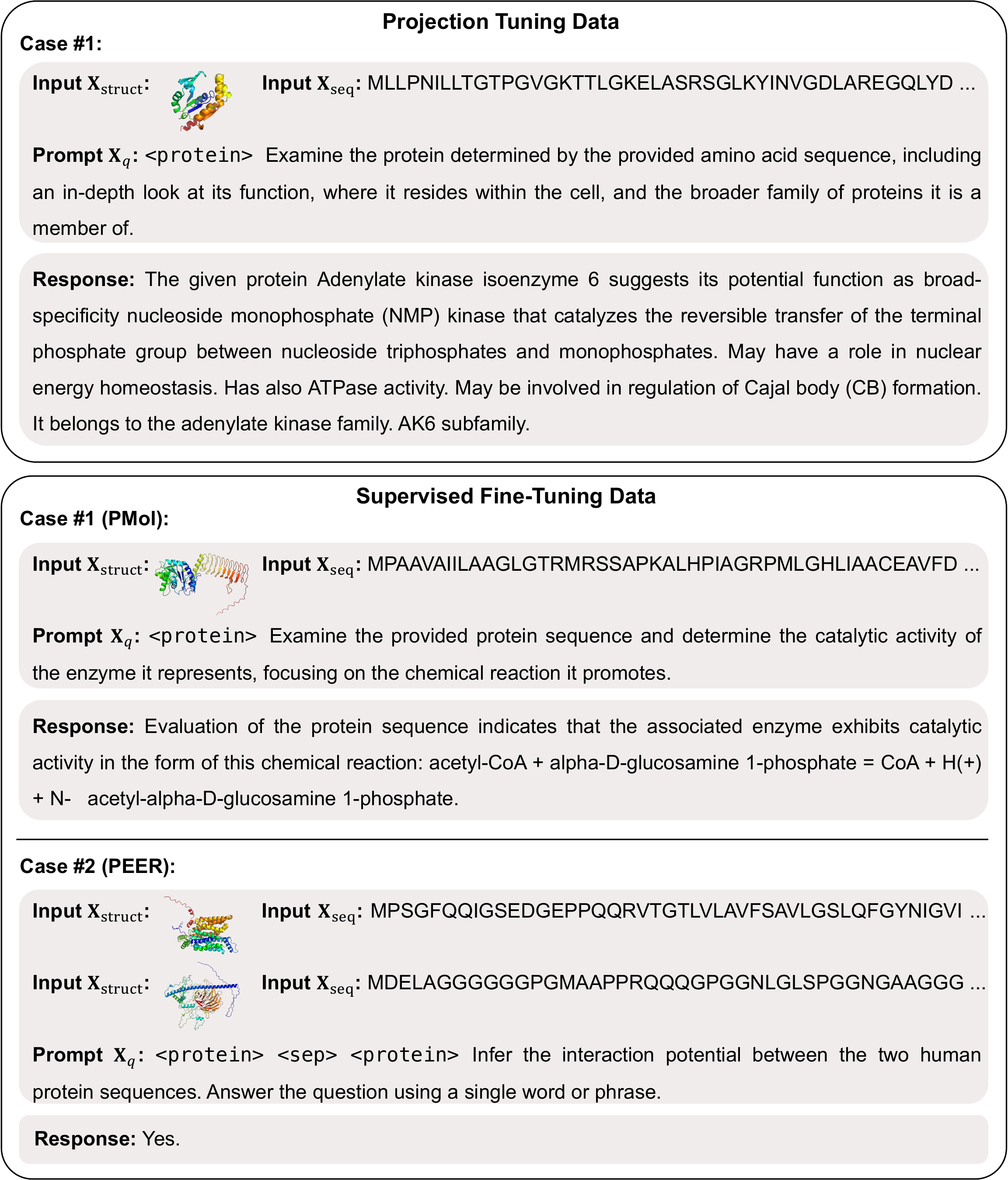}
    \caption{An example of the projection tuning data and supervised fine-tuning data. Note that the special token \texttt{<protein>} denotes the fused protein representations of structural and sequential features.}
    \label{fig:data-example}
\end{figure}

In this section, we introduce the construction of protein instruction-following data.
It consists of two sets, projection tuning and supervised fine-tuning, which are used at different training stages, described in Sec.~\ref{sec:approach}.
An example of the projection tuning data and supervised fine-tuning data is illustrated in Fig.~\ref{fig:data-example}.
Formally, for each example, we define $\mathbf{X}_p, \mathbf{X}_q, \mathbf{X}_a$ as the protein (fused protein representations, consisting of the 3D structure \(\mathbf{X}_{\text{struct}}\) and the amino acid sequence \(\mathbf{X}_{\text{seq}}\)), the protein-related natural language question, and the corresponding answer, respectively.

\paragraph{Projection Tuning Data}
It consists of protein-text pairs originated from the Swiss-Prot \citep{uniprot2023uniprot} database. Due to limited computational resources, we directly utilize the 3D structures predicted by AlphaFold-2 \citep{jumper2021highly} from Swiss-Prot.
The database contains 571K manually-annotated records, each containing information including protein name, subcellular location, function and families. 
For $\mathbf{X}_q$, we construct 10 templates that ask the model to briefly describe the input protein $\mathbf{X}_p$ from various aspects.
For $\mathbf{X}_a$, information is extracted from the filtered Swiss-Prot annotation and constructed using a pre-defined template to ensure the consistency and clarity of protein descriptions. 
The question and answer templates are listed in Fig.~\ref{fig:template-pretrain}.

\paragraph{Supervised Fine-tuning Data}
To align the model to follow a variety of instructions, we present and curate diverse instruction-following data about the provided proteins, by verbalizing protein-related tasks.
It consists of 10 tasks including Mol-Instructions \citep{fang2023mol} and PEER benchmark \citep{xu2022peer}. We use ESMFold \citep{lin2022language} to accelerate protein structure prediction for sequences in these two datasets.
\begin{itemize}
    \item \textbf{Mol-Instructions} is a comprehensive instruction dataset designed for the biomolecular realm.
    It includes three components: molecule-oriented instructions, protein-oriented instructions and biomolecular text instructions.
    We adopt protein-oriented instructions in Mol-Instructions (named PMol) for the supervised fine-tuning stage.
    PMol details will be discussed in Sec.~\ref{sec:molinst}.
    For each task in Mol-Instructions, we make simple modifications to the original prompts to fit our use cases and ensure coherence.
    Details are discussed in Appendix~\ref{sec:data-finetune} and some modification cases are listed in Fig.~\ref{fig:molinst-template}.
    \item \textbf{PEER} is a comprehensive benchmark for general protein sequence understanding tasks including protein localization prediction, protein structure prediction and protein-protein interaction prediction.
    PEER benchmark details will be discussed in Sec.~\ref{sec:peer}.
    For each task in PEER benchmark, there are 10 question templates and 1 answer template, some of which are listed in Fig.~\ref{fig:peer-template}. 
    In response templates for other tasks, categories are represented by natural language. 
    However, for fold classification, we use integers from 0 to 1,194 due to the large number of categories.
\end{itemize}

%% file: docs/030experiment.tex
We evaluate \mymodel \footnote{Unless specified otherwise, \mymodel refers to \mympnnmodel.} on downstream tasks including protein understanding tasks based on Mol-Instructions (Sec.~\ref{sec:molinst}) and protein property prediction tasks based on PEER benchmark (Sec.~\ref{sec:peer}). Additionally, the structure encoder in our approach is replaced with GearNet to construct \mygearmodel for the experiments, and further experiments on the substitution of protein sequence encoders are provided in Appendix~\ref{sec:more-eval}. We evaluate our approach in both zero-shot settings, where only the projection layers are aligned during the projection tuning stage, and in supervised fine-tuning without the projection tuning stage. Details of the experimental setups are discussed in Appendix~\ref{sec:exp-setup}.

\subsection{Protein Understanding Tasks}
\label{sec:molinst}
\paragraph{Task Descriptions}
Protein understanding tasks use PMol for fine-tuning and evaluation, which consist of four distinct tasks with datasets constructed based on UniProtKB \citep{uniprot2021uniprot}. Protein function prediction outputs the function of the given protein. Catalytic activity prediction outputs the catalytic activity of the input protein and the chemical reactions it promotes. Domain/motif prediction outputs the domains or motifs that the given protein may contain. Functional description generation outputs the description of the input protein's function, subcellular localization, and any biological processes it may be a part of.

\paragraph{Baselines}
We compare our approach with the protein-oriented LLMs in Mol-Instructions including LLaMA \citep{touvron2023llama}, Alpaca \citep{tloen2023alpaca}, Baize \citep{xu2023baize}, ChatGLM \citep{zeng2022glm}, Galactica \citep{taylor2022galactica} and Vicuna. Apart from these LLMs, we use Prot2Text \citep{abdine2024prot2text} and ProLLaMA \citep{lv2024prollama} as our baseline models in zero-shot settings. These models lack support for arbitrary prompts. Prot2Text is designed to generate protein descriptions, while ProLLaMA predicts protein superfamilies. Therefore, we evaluate protein function prediction for Prot2Text and domain/motif prediction for ProLLaMA. For protein understanding tasks, we follow Mol-Instructions, taking ROUGE-L \citep{lin2004rouge} as the evaluation metric. Details of ROUGE-L implementation are discussed in Appendix~\ref{sec:eval-impl}.

\input{tables/tab-molinst}

\paragraph{Results}
As shown in Tab.~\ref{tab:molinst}, \mymodel and \mygearmodel with zero-shot not only handle all protein understanding tasks but also outperform Prot2Text and ProLLaMA. Furthermore, they surpass or approach ChatGLM, Llama-2-7B-Chat and Vicuna fine-tuned on protein-oriented instructions by 1\%-8\%, demonstrating that during the projection tuning stage, \mymodel and \mygearmodel learn protein knowledge and can follow human instructions to generalize their knowledge for various downstream tasks. Additionally, \mymodel outperforms all baseline models, including Llama-2-7B-Chat fine-tuned on the complete Mol-Instructions dataset (Mol) after supervised fine-tuning, highlighting the effectiveness of our approach. Notably, our approach uses a relatively small amount of data and has significantly fewer trainable parameters than the baseline models using full-parameter fine-tuning. The experimental results highlight the importance of leveraging the multimodal structure and sequence representations during training LLMs.

\subsection{Protein Property Prediction Tasks}
\label{sec:peer}
\paragraph{Task Descriptions}
Protein property prediction tasks use PEER benchmark for fine-tuning and evaluation, which consist of 6 tasks. Solubility prediction \citep{khurana2018deepsol}, defined as a binary classification task, aims to predict whether a given protein is soluble or not. Subcellular localization prediction \citep{almagro2017deeploc}, defined as a ten-class classification task, aims to predict where a given protein locates in the cell. Binary localization prediction, a simplified version of subcellular localization prediction, is defined as a binary classification task that aims to determine whether a given protein is soluble or membrane-bound. Fold classification \citep{fox2014scope,hou2018deepsf} requires the model to classify the global structural topology of a given protein into one of 1195 classes at the fold level. Yeast PPI prediction \citep{guo2008using} and human PPI prediction \cite{peri2003development,pan2010large} are defined as binary localization tasks, which aim to predict whether two given yeast or human proteins interact or not respectively. It is worth noting that for all tasks, protein sequences in the training set with high similarity to those in the test set are excluded based on the sequence identity. For example, sequences with \(\geq\) 30\% identity are excluded in the solubility prediction task. Therefore, a key challenge in protein property prediction tasks lies in evaluating a model's ability to generalize across dissimilar protein sequences.

\paragraph{Baselines}
We compare our approach with the following baselines in PEER benchmark. Feature engineers include Dipeptide Deviation from Expected Mean (DDE) \citep{saravanan2015harnessing} and Moran correlation (Moran) \cite{feng2000prediction}. Protein sequence encoders include LSTM \citep{hochreiter1997long}, Transformer \citep{vaswani2017attention}, shallow CNN \citep{shanehsazzadeh2020transfer} and ResNet \citep{he2016deep}. Pre-trained PLMs include ProtBert \citep{elnaggar2021prottrans} and ESM-1b \cite{rives2021biological}. Protein-oriented LLMs include Llama-3-8B-Instruct \citep{dubey2024llama3} and InstructProtein \citep{wang2024instructprotein}. For protein property prediction tasks, we take accuracy (Acc) as the evaluation metric.

\input{tables/tab-peer}

\paragraph{Results}
As displayed in Tab.~\ref{tab:peer}, \mymodel with zero-shot achieves a comparable performance with both task-specific and single models across several tasks including solubility prediction, binary localization prediction and PPI prediction tasks. \mygearmodel with zero-shot performs relatively worse than \mymodel due to its lower capability to follow instructions. Besides, the supervised fine-tuning stage significantly enhances the performance of our approach, enabling it to outperform or approach the previous state-of-the-art models including ProtBert, ESM-1b and InstructProtein across multiple tasks. Additionally, we add a linear classification head to ESM-2 and fine-tune it with full parameters for 1K steps per task. ESM-2 achieves an average score of 67.10, demonstrating that ESM-2 is a competitive baseline compared to ESM-1b. Our approach surpasses ESM-2 on binary localization prediction and human PPI prediction tasks while approaching it on other tasks. We find that when jointly fine-tuning models on various tasks through multi-task learning, the performance improves compared to task-specific models, demonstrating no negative impact between tasks. However, more interrelated downstream tasks or data could be introduced in future work to further enhance the model's performance.

In particular, compared to task-specific models, \mymodel outperforms or approaches them across several tasks, except for fold classification. A possible explanation for the promising results is that \mymodel learns the properties of proteins through various tasks by leveraging multimodal structure and sequence representations. Notably, both \mymodel and Llama-3-8B-Instruct achieve the best performance on binary localization task, demonstrating the significant potential of LLMs to understand proteins. Additionally, BioT5 and BioT5+ are trained to solve four tasks, excluding subcellular localization prediction and fold classification, achieving average accuracy scores of 79.36 and 79.01, respectively, across these tasks. In comparison, \mymodel has an average score of 75.57, showing comparable performance with these two models. A possible explanation for the relatively higher accuracy of these two models is that they incorporate molecules as a new modality and leverage more molecule-related data, potentially yielding positive effects, while we focus only on protein structures and sequences. A similar result can be observed in Tab.~\ref{tab:molinst}, where Llama-2-7B-Chat fine-tuned on the complete Mol-Instructions dataset, including molecule-related data, performs significantly better than when fine-tuned only on protein-oriented instructions. We leave the exploration of incorporating more biomolecular modalities, such as molecules and DNA, for future work.

Additionally, compared to Llama-3-8B-Instruct, used as a text decoder in our approach, \mymodel improves performance on all tasks by incorporating the multimodal structure and sequence representations of proteins.

\subsection{Ablation Study}
\label{sec:ablation}

In this section, we conduct ablation studies to explore the effect of the projection tuning stage, structure and sequence representations, and the fusion method.
For a fair comparison, the baselines are trained using the same experimental setups discussed in Appendix~\ref{sec:exp-setup} with 10K steps. 
We conduct more evaluations on protein property prediction tasks in Appendix~\ref{sec:more-eval}, where we further discuss the effect of protein sequence encoder sizes.

\paragraph{Effect of the projection tuning stage}

We compare the performance of \mymodel with and without the projection tuning stage in Tab.~\ref{tab:ablation-pt-more}. \mymodel, when continued to be fine-tuned after the projection tuning stage, achieves lower performance compared to direct supervised fine-tuning. A possible reason is that we use protein structures predicted by AlphaFold-2 during the projection tuning stage, while structures predicted by ESMFold are used during the supervised fine-tuning stage. The zero-shot ability discussed in Sec.~\ref{sec:molinst} and Sec.~\ref{sec:peer} highlights the effectiveness of the projection tuning stage, demonstrating that the model learns to generalize its structural knowledge from structures predicted by AlphaFold-2 to those predicted by ESMFold during inference. However, it's challenging for models to learn the gap between protein structures predicted by AlphaFold-2 and ESMFold when the structure encoder and projection layer are updated simultaneously during the supervised fine-tuning stage.

\input{tables/tab-ablation-pt-more}

\paragraph{Effect of structure and sequence representations}

As shown in Tab.~\ref{tab:ablation-features-more}, \mymodel incorporating both structure and sequence representations outperforms those utilizing either alone, demonstrating the effectiveness of integrating both protein representations. \mygearmodel significantly enhances the performance by leveraging structure and sequence representations. Furthermore, \mymodel with only a sequence-based protein encoder surpasses the one with only a structure-based protein encoder, regardless of the protein structure encoder chosen. A possible explanation is that the features extracted by ESM-2 implicitly contain structural information, indicating that sequence representations are easier for LLMs to learn. Moreover, the parameters of the structure encoder are one to two orders of magnitude fewer than those of the sequence encoder, leading to a more limited extraction of structural features. A case study on the effect of structure and sequence representations is conducted in Appendix~\ref{sec:case-study}.

\input{tables/tab-ablation-features-more}

\paragraph{Effect of the fusion method}

To evaluate the effect of the fusion method, we directly use the structure and sequence features instead of fusing their representations. As shown in Tab.~\ref{tab:ablation-fusion-more}, \mymodel with fused representations surpass the one without, demonstrating the effectiveness of our fusion method. However, \mygearmodel without fused representations outperforms the one with them, indicating that for different protein structure encoders, different fusion methods may be chosen. Furthermore, the fused representations reduce the token cost of the LLM, resulting in approximately 20\% lower inference latency.

\input{tables/tab-ablation-fusion-more}

%% file: tables/tab-molinst.tex
\begin{table*}[!htb]
    \centering
    \resizebox{\textwidth}{!}{%
    \begin{tabular}{lccccccc}
    \toprule
    \multirow{2}{*}{{Models}}  & \multirow{2}{*}{{Data}} & {Trainable/Total} & \multicolumn{5}{c}{ROUGE-L($\uparrow$)} \\ 
    \cmidrule[0.5pt](lr){4-8} 
     & &{\#Params.} & PF & GF & CA & DP & Avg. \\ \midrule
     \multicolumn{8}{l}{~ ~ \emph{Models in zero-shot settings}} \\
     Galactica & - & - & 0.12 & 0.12 & 0.13 & 0.09 & 0.1150 \\
     Prot2Text & -  & - & 0.14 & - & - & - & 0.1400 \\
     ProLLaMA & -  & - & - & - & - & 0.02 & 0.0200 \\
     \chl \mygearmodel & \chl  -  &\chl - & \chl 0.16 & \chl 0.16 & \chl 0.21 & \chl 0.15 & \chl 0.1700 \\
     \chl \mympnnmodel  & \chl  -  &\chl - & \chl 0.16  & \chl 0.14 & \chl 0.15 & \chl 0.11 & \chl 0.1400 \\ \midrule
     \multicolumn{8}{l}{~ ~ \emph{Fine-tuned models}} \\
     ChatGLM & PMol &6B/6B &0.15 & 0.14 & 0.13 & 0.10 & 0.1300 \\
     Llama-2-7B-Chat & PMol & 7B/7B&0.15 & 0.14 & 0.16 & 0.12 & 0.1425 \\
     Llama-2-7B-Chat & Mol & 7B/7B & \underline{0.42} & \underline{0.44} & \underline{0.52} & \underline{0.46} & \underline{0.4600} \\ 
     Vicuna & PMol & 7B/7B&0.07 & 0.08 & 0.08 & 0.06 & 0.0725 \\
     Alpaca & PMol & 7B/7B &0.2 & 0.1 & 0.23 & 0.12 & 0.1625 \\
     Baize & PMol & 7B/7B &0.2 & 0.15 & 0.22 & 0.13 & 0.1750 \\
     \chl \mygearmodel & \chl  PMol, PEER  &\chl 720M/8.8B (8.2\%) & \chl \textit{0.25} & \chl \textit{0.32} & \chl \textit{0.34} & \chl \textit{0.31} & \chl \textit{0.3050} \\
     \chl \mympnnmodel & \chl PMol, PEER & \chl 690M/8.8B (7.9\%) &\chl \textbf{0.48} & \chl \textbf{0.50}  & \chl \textbf{0.60} & \chl \textbf{0.50} & \chl \textbf{0.5200} \\
     \bottomrule
    \end{tabular}%
    }
    \caption{Results of protein understanding tasks (\textbf{Best}, \underline{Second Best}, \textit{Third Best}). PF refers to protein function prediction. GF refers to functional description generation. CA refers to catalytic activity prediction. DP refers to domain/motif prediction. Note that Mol refers to the Mol-Instructions with 3 components: molecule-oriented instructions, protein-oriented instructions (named PMol), and biomolecular text instructions. - indicates the data is not applicable to the task.}
    \label{tab:molinst}
\end{table*}

%% file: tables/tab-peer.tex
\begin{table*}[!t]
  \centering
  \resizebox{\textwidth}{!}{%
  \begin{tabular}{@{}lccccccc@{}}
    \toprule
    Models & Sol & Sub & Bin & Fold & Yst & Hum & Avg. \\ \midrule
     \multicolumn{8}{l}{~ ~ \emph{Task-specific models}} \\
    DDE & 59.77\(\pm\)1.21 & 49.17\(\pm\)0.40 & 77.43\(\pm\)0.42 & 9.57\(\pm\)0.46 & 55.83\(\pm\)3.13 & 62.77\(\pm\)2.30 & 52.42 \\
    Moran & 57.73\(\pm\)1.33  & 31.13\(\pm\)0.47 & 55.63\(\pm\)0.85 & 7.10\(\pm\)0.56 & 53.00\(\pm\)0.50 & 54.67\(\pm\)4.43 & 43.21 \\
    \cmidrule{2-8}
    CNN & 64.43\(\pm\)0.25  & 58.73\(\pm\)1.05 & 82.67\(\pm\)0.32 & 10.93\(\pm\)0.35 & 55.07\(\pm\)0.02 & 62.60\(\pm\)1.67 & 55.74 \\
    ResNet & 67.33\(\pm\)1.46  & 52.30\(\pm\)3.51 & 78.99\(\pm\)4.41 & 8.89\(\pm\)1.45 & 48.91\(\pm\)1.78 & 68.61\(\pm\)3.78 & 54.17 \\
    LSTM & \textit{70.18\(\pm\)0.63}  & 62.98\(\pm\)0.37 & 88.11\(\pm\)0.14 & 8.24\(\pm\)1.61 & 53.62\(\pm\)2.72 & 63.75\(\pm\)5.12 & 57.81 \\
    Transformer & 70.12\(\pm\)0.31  & 56.02\(\pm\)0.82 & 75.74\(\pm\)0.74 & 8.52\(\pm\)0.63 & 54.12\(\pm\)1.27 & 59.58\(\pm\)2.09 & 54.02 \\ 
    \cmidrule{2-8}
    ProtBert & 68.15\(\pm\)0.92 & \underline{76.53\(\pm\)0.93} & 91.32\(\pm\)0.89 & \underline{16.94\(\pm\)0.42} & \textbf{63.72\(\pm\)2.80} & \underline{77.32\(\pm\)1.10} & \underline{65.66} \\
    ESM-1b & \underline{70.23\(\pm\)0.75}  & \textbf{78.13\(\pm\)0.49} & 92.40\(\pm\)0.35 & \textbf{28.17\(\pm\)2.05} & \textit{57.00\(\pm\)6.38} & \textbf{78.17\(\pm\)2.91} & \textbf{67.35} \\ \midrule
     \multicolumn{8}{l}{~ ~ \emph{Single models in zero-shot settings}} \\
    \chl \mygearmodel & \chl 34.65\(\pm\)0.48 & \chl - & \chl 50.89\(\pm\)0.88 & \chl - & \chl 2.52\(\pm\)0.11 & \chl 3.70\(\pm\)0.10 & \chl 22.94 \\ 
    \chl \mympnnmodel & \chl 50.76\(\pm\)0.47 & \chl 8.16\(\pm\)0.27 & \chl \textit{92.85\(\pm\)0.21} & \chl 0.65\(\pm\)0.06 & \chl 53.59\(\pm\)0.36 & \chl 50.21\(\pm\)0.79 & \chl 42.70 \\ \midrule
    \multicolumn{8}{l}{~ ~ \emph{Single fine-tuned models}} \\
    InstructProtein & 69.08\(\pm\)0.00 & 70.79\(\pm\)0.00 & 85.19\(\pm\)0.00 & 10.86\(\pm\)0.00 & - & - & 58.98 \\
    Llama-3-8B-Instruct & 69.13\(\pm\)0.39 & 51.36\(\pm\)0.06 & \underline{98.91\(\pm\)0.00} & 8.77\(\pm\)0.40 & 56.05\(\pm\)0.53 & 62.87\(\pm\)0.34 & 57.85 \\
    \chl \mygearmodel & \chl 61.13\(\pm\)0.15 & \chl 42.27\(\pm\)0.23 & \chl 85.21\(\pm\)0.34 & \chl 3.11\(\pm\)0.36 & \chl 50.42\(\pm\)0.57 & \chl 67.72\(\pm\)1.08 & \chl 51.64 \\
    \chl \mympnnmodel & \chl \textbf{72.37\(\pm\)0.35} & \chl \textit{73.25\(\pm\)0.27} & \chl \textbf{99.73\(\pm\)0.05} & \chl \textit{10.96\(\pm\)0.36} & \chl \underline{57.45\(\pm\)0.78} & \chl \textit{72.71\(\pm\)1.15} & \chl \textit{64.41} \\
    \bottomrule
  \end{tabular}%
  }
  \caption{Results (in \%) of protein property prediction tasks (\textbf{Best}, \underline{Second Best}, \textit{Third Best}). Sol represents solubility prediction. Sub represents subcellular localization prediction. Bin represents binary localization prediction. Fold represents fold classification. Yst represents yeast PPI prediction. Hum represents human PPI prediction. - indicates the data is not applicable to the task.}
  \label{tab:peer}
\end{table*}

%% file: tables/tab-ablation-pt-more.tex
\begin{table}[!htb]
    \centering
    \resizebox{\linewidth}{!}{%
    \begin{tabular}{lccccc}
    \toprule
        Models & PF & GF & CA & DP & Avg. \\ \midrule
        \mymodel & 0.43 & 0.46 & 0.55 & 0.48 & \textbf{0.4800} \\
        \begin{tabular}[c]{@{}l@{}}\mymodel \\ ~ ~ w/ PT \end{tabular} & 0.28 & 0.34 & 0.40 & 0.34 & 0.3400 \\ \bottomrule
    \end{tabular}%
    }
    \caption{Effect of the projection tuning stage. The experiments are conducted on protein understanding tasks. PT refers to the projection tuning stage.}
    \label{tab:ablation-pt-more}
\end{table}

%% file: tables/tab-ablation-features-more.tex
\begin{table}[!htb]
    \centering
    \resizebox{\linewidth}{!}{%
    \begin{tabular}{lccccc}
    \toprule
        Models & PF & GF & CA & DP & Avg. \\ \midrule
        \mymodel & 0.43 & 0.46 & 0.55 & 0.48 & \textbf{0.4800} \\
        \begin{tabular}[c]{@{}l@{}}\mymodel \\ ~ ~ w/o ProteinMPNN \end{tabular} & 0.39 & 0.46 & 0.54 & 0.48 & 0.4675 \\
        \begin{tabular}[c]{@{}l@{}}\mymodel \\ ~ ~ w/o ESM-2 \end{tabular} & 0.35 & 0.42 & 0.50 & 0.41 & 0.4200 \\ \midrule
        \mygearmodel & 0.30 & 0.32 & 0.36 & 0.30 & \textbf{0.3200} \\ \\
        \begin{tabular}[c]{@{}l@{}}\mygearmodel \\ ~ ~ w/o ESM-2 \end{tabular} & 0.20 & 0.28 & 0.31 & 0.29 & 0.2700 \\ \bottomrule
    \end{tabular}%
    }
    \caption{Effect of structure and sequence representations. The experiments are conducted on protein understanding tasks. Note that \mygearmodel (w/o GearNet) is equivalent to \mymodel (w/o ProteinMPNN), as both only utilize the sequence representations extracted by the protein sequence encoder.}
    \label{tab:ablation-features-more}
\end{table}

%% file: tables/tab-ablation-fusion-more.tex
\begin{table}[!htb]
    \centering
    \resizebox{\linewidth}{!}{%
    \begin{tabular}{lcccccc}
    \toprule
        Models & PF & GF & CA & DP & Avg. & Latency \\ \midrule
        \mymodel & 0.43 & 0.46 & 0.55 & 0.48 & \textbf{0.4800} & \textbf{\(\times\)1.0} \\
        \begin{tabular}[c]{@{}l@{}}\mymodel \\ ~ ~ w/o fused representations \end{tabular} & 0.42 & 0.44 & 0.55 & 0.48 & 0.4725 & \(\times\)1.17 \\ \midrule
        \mygearmodel & 0.30 & 0.32 & 0.36 & 0.30 & 0.3200 & \textbf{\(\times\)0.8} \\
        \begin{tabular}[c]{@{}l@{}}\mygearmodel \\ ~ ~ w/o fused representations \end{tabular} & 0.31 & 0.34 & 0.39 & 0.36 & \textbf{0.3500} & \(\times\)1.0  \\ \midrule
    \end{tabular}%
    }
    \caption{Effect of the fusion method. The experiments are conducted on protein understanding tasks.}
    \label{tab:ablation-fusion-more}
\end{table}

%% file: docs/050conclusion.tex
In this paper, we propose \mymodel, a multimodal framework that connects ProteinMPNN, ESM-2 650M and Llama-3 8B for protein understanding through a two-stage training process. 
Experiments demonstrate that after the projection tuning stage, \mymodel in zero-shot settings outperforms the fine-tuned baselines with full parameters, surpassing the current state-of-the-art model with supervised fine-tuning on the Mol-Instructions. Additionally, our approach achieves promising results that are competitive with state-of-the-art task-specific baselines on the PEER benchmark.

%% file: docs/055limitations.tex
\mymodel incorporates the structure and sequence representations of proteins to enhance LLM's understanding of proteins. Due to the lack of experimentally determined structures for many proteins in our experiments, we use 3D structures predicted by AlphaFold-2 and ESMFold to fully leverage the data. These computationally predicted structures generally have relatively lower accuracy compared to wet lab experiments. Besides, training a single model to predict various protein properties presents challenges, causing \mymodel to only approach the state-of-the-art performance for some tasks in protein property prediction.

%% file: docs/060appendix.tex
\section{Further Related Work}
\label{sec:related-more}

\paragraph{Large Language Models}
Llama-3 \citep{dubey2024llama3} is a family of pre-trained and fine-tuned open-source LLMs with sizes ranging from 8B to 70B parameters. 
Baize \citep{xu2023baize} is an open source model aligned by introducing self-distillation with feedback.
ChatGLM \citep{zeng2022glm} is an open bilingual language model based on General Language Model (GLM) framework \citep{du2021glm}, with 6B parameters.
Galactica \citep{taylor2022galactica} is a large language model that trained on an extensive scientific corpus of papers and knowledge bases, capable of storing, combining and reasoning about scientific knowledge.

\paragraph{Vision Language Models}
Vision Language Models (VLMs) such as LLaVA \cite{liu2024visual} and InstructBLIP \cite{dai2024instructblip} have demonstrated their capability to understand visual content.
LLaVA introduces an end-to-end LMM that connects a visual encoder and a LLM for visual and language understanding. InstructBLIP proposes a instruction tuning framework towards generalized vision-language models.
Geneverse \citep{liu2024geneverse} is a collection of fine-tuned LLMs and VLMs for three novel tasks in genomic and proteomic research. For the protein task, Geneverse uses protein structure images with the fixed capture angle. In contrast, our approach treats amino acid sequences and 3D structures as distinct modalities. Unlike the visual information provided by protein structure images, \mymodel captures sequential and structural features from the primary and tertiary structures of proteins, offering a complementary perspective on multimodal protein representations.

\paragraph{Protein-oriented LLMs}
OntoProtein \citep{zhang2022ontoprotein} integrate external factual knowledge from gene ontology into PLMs to enhance protein representations.
ProteinChat \citep{guo2023proteinchat} utilizes a Graph Neural Network encoder block combined with a Transformer encoder block to extract features from protein structures. Instead of introducing complex Transformer blocks, \mymodel uses MLP as a lightweight and cost-effective approach to align different modalities.
ProtST \citep{xu2023protst} is a framework to enhance protein sequence understanding through biomedical texts by utilizing a Protein Language Model (PLM) and a Biomedical Language Model (BLM). The weights of BLM are initialized from PubMedBERT-abs \citep{gu2021domain}, which is pre-trained on PubMed abstracts.
ProtChatGPT \citep{wang2024protchatgpt} is trained with sequence-text pairs using a Protein-Language Pretraining Transformer initialized with the pre-trained weights of PubMedBERT \citep{gu2021domain} to incorporate external knowledge. Compared to ProtST and ProtChatGPT, \mymodel does not rely on LLMs specifically trained on external biomedical domain knowledge. Instead, we tune the projection layers from scratch, highlighting the generalizability of our plug-and-play architecture.
ProteinGPT \citep{xiao2024proteingpt} is a multimodal protein chat system trained through a two-stage process: modality alignment and instruction tuning. In contrast, our approach demonstrates that the initial alignment stage can be optional (Sec.~\ref{sec:ablation}), making our method more efficient and reducing training costs. Furthermore, \mymodel employs a template-based strategy to construct the instruction-following dataset, avoiding the use of GPT-4o for generating the QA dataset as described in \citet{xiao2024proteingpt}. This template-driven approach not only enables zero-shot capability in handling unseen instructions during the projection tuning stage (Sec.~\ref{sec:molinst} and Sec.~\ref{sec:peer}), showcasing the generalization and robustness of our method, but also eliminates the API call costs.
ProteinCLIP \citep{wu2024proteinclip} performs contrastive learning between protein sequences and texts by employing a frozen protein encoder and a frozen text embedding encoder. While ProteinCLIP is designed for protein function related tasks, our approach can be adapted to various downstream tasks including catalytic activity prediction and domain/motif prediction.
ProtLLM \citep{zhuo2024protllm} is a cross-modal LLM designed for protein-centric and protein-language tasks. Unlike ProtLLM, our approach integrates both structure and sequence representations of proteins, rather than relying solely on the sequence modality. This allows for a more comprehensive understanding of protein features.

\paragraph{Protein Representations}
DDE \citep{saravanan2015harnessing}, based on the dipeptide frequency within the protein sequence, and Moran \citep{feng2000prediction}, which defines the distribution of amino acid properties along a protein sequence, are two typical protein sequence feature descriptors.
Shallow CNN \citep{shanehsazzadeh2020transfer} and ResNet \citep{he2016deep} are protein sequence encoders designed to capture the short-range interactions within the protein sequence, while LSTM \citep{hochreiter1997long} and Transformer \citep{vaswani2017attention} aim to capture the long-range interactions. The output layers of these protein sequence encoders aggregate the representations of different residues into a protein-level representation.
Apart from these methods, some recent work has focused on simultaneously encoding protein sequences and structures.
SaProt \citep{su2023saprot} integrates residue tokens with structure tokens and is pre-trained on approximately 40M sequences and structures. In contrast, \mymodel avoids the introduction of additional tokens and instead aligns the structure and sequence representations of proteins with embeddings from natural language prompts, achieving this with significantly fewer sequences and structures. ESM-3 \citep{hayes2024simulating} is a generative language model that reasons over the sequence, structure, and function of proteins. Compared to ESM-3, our approach treats arbitrary protein functions as natural language rather than discrete function tokens or keywords, showing flexibility in both understanding and generating descriptive texts about protein functions.

\section{Dataset Construction Details}
\label{sec:data-construct}

\subsection{Projection Tuning Data}
\label{sec:data-pretrain}

\begin{figure}[htb]
    \centering
    \includegraphics[width=0.945\linewidth]{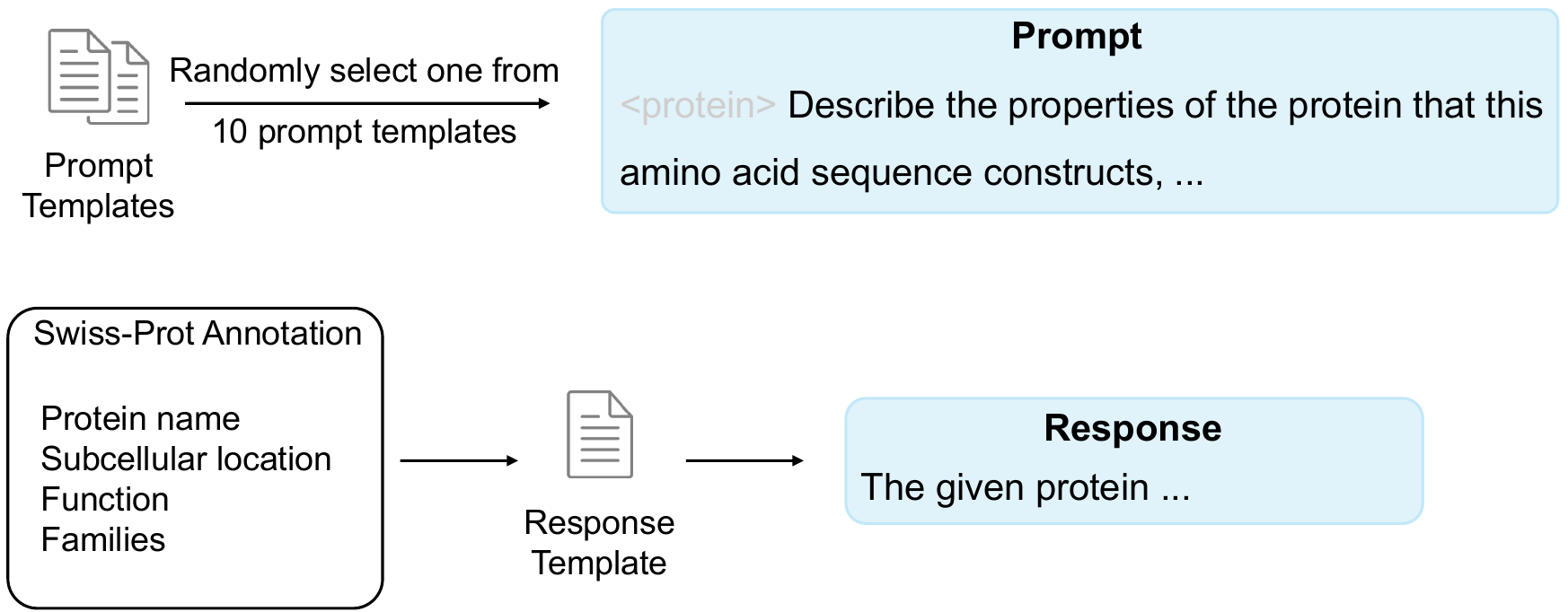}
    \caption{Overview of the projection tuning data construction.}
    \label{fig:data-pretrain}
\end{figure}

\begin{figure*}[!ht]
    \centering
    \includegraphics[width=\linewidth]{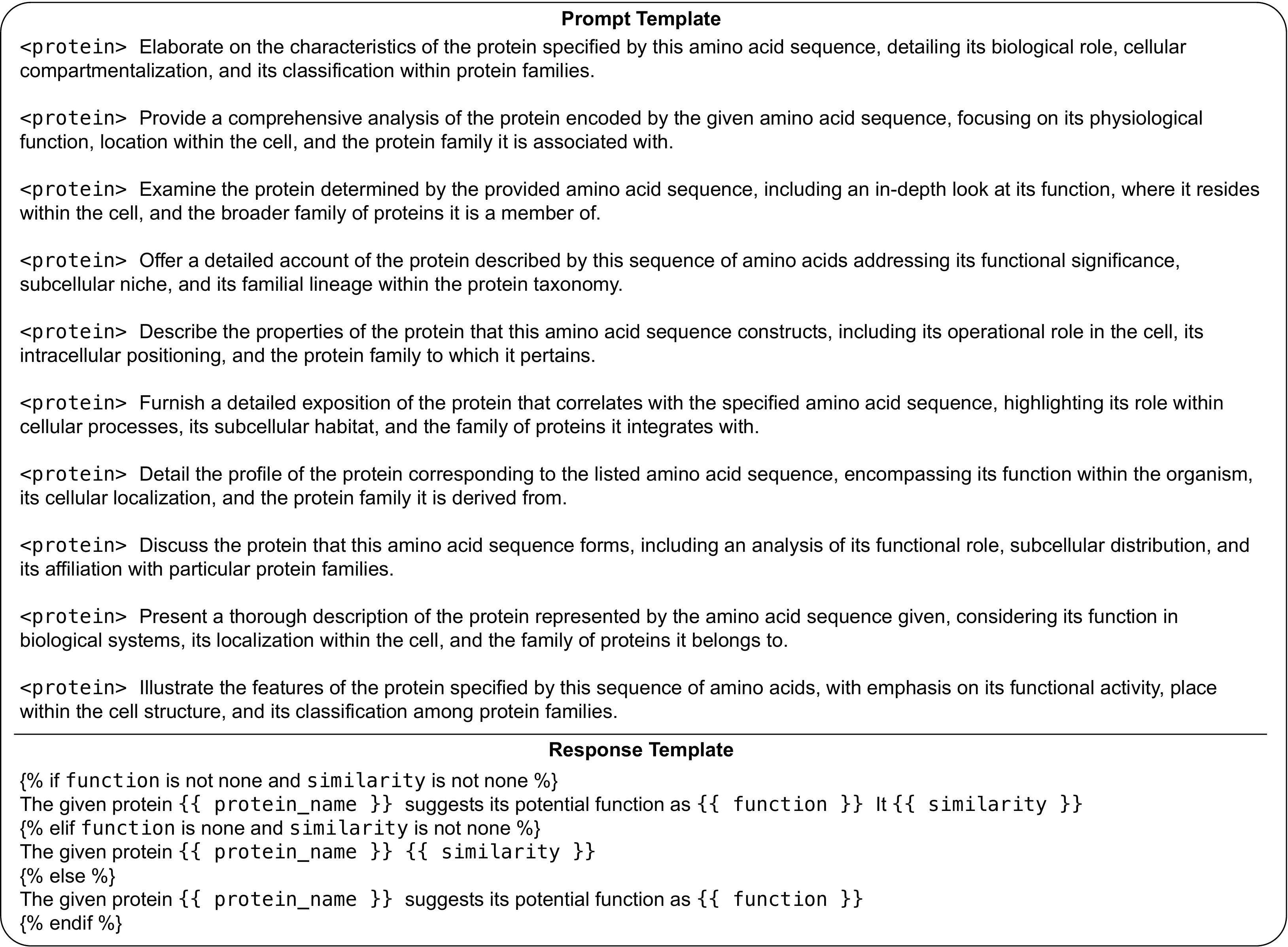}
    \caption{The prompt and response template of the projection tuning data. In the response template, \texttt{similarity} refers to the families of the protein in Swiss-Prot.}
    \label{fig:template-pretrain}
\end{figure*}

As illustrated in Fig.~\ref{fig:data-pretrain}, projection tuning data consists of sequence-description pairs originated from the Swiss-Prot \citep{uniprot2023uniprot} database. The database contains 571K manually-annotated records, each containing information including protein name, subcellular location, function and families. To prevent from data leakage, we filter the Swiss-Prot annotation to 369K as our projection tuning data based on the downstream tasks.

For prompts, we construct 10 templates that ask the model to briefly describe the input protein from various aspects. For responses, information is extracted from the filtered Swiss-Prot annotation and constructed using a pre-defined template to ensure the consistency and clarity of protein descriptions. The prompt and response templates are listed in Fig.~\ref{fig:template-pretrain}.

\subsection{Supervised Fine-Tuning Data}
\label{sec:data-finetune}
\begin{figure}[!h]
    \centering
    \includegraphics[width=\linewidth]{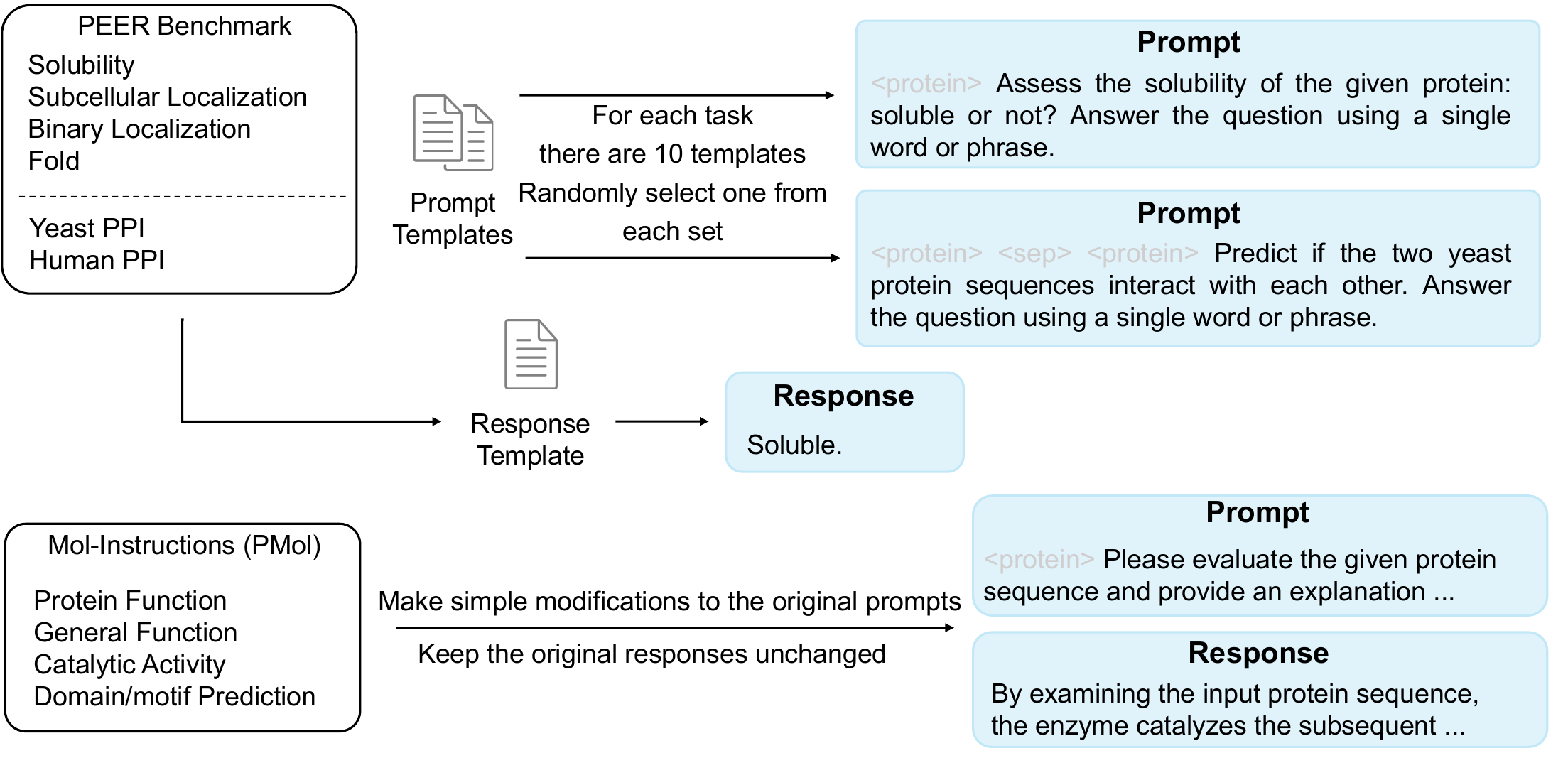}
    \caption{Overview of the supervised fine-tuning data construction.}
    \label{fig:data-finetune}
\end{figure}

\begin{figure*}[!htb]
    \centering
    \includegraphics[width=\textwidth]{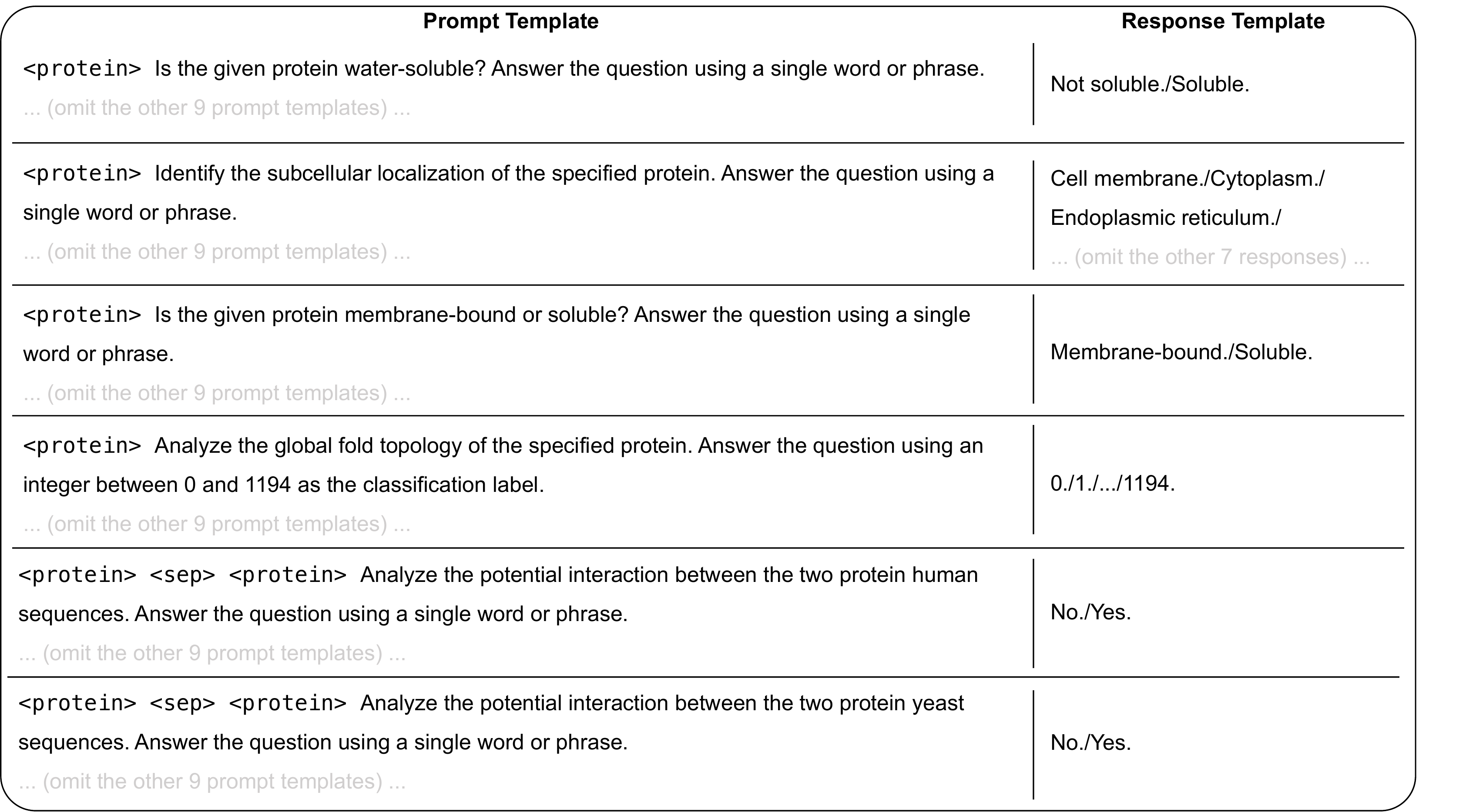}
    \caption{The prompt and response template of PEER benchmark in the supervised fine-tuning data.}
    \label{fig:peer-template}
\end{figure*}

As shown in Fig.~\ref{fig:data-finetune}, fine-tuning dataset consists of 10 tasks including PEER benchmark \citep{xu2022peer} and Mol-Instructions \citep{fang2023mol}. For each task in PEER benchmark, there are 10 prompt templates and 1 response template, some of which are listed in Fig.~\ref{fig:peer-template}. Except for fold classification, the categories in the response templates for other tasks are represented by natural language. For fold classification, we use integers ranging from 0 to 1194 to represent its categories due to the excessive number of categories.

\begin{figure*}[!htb]
    \centering
    \includegraphics[width=\textwidth]{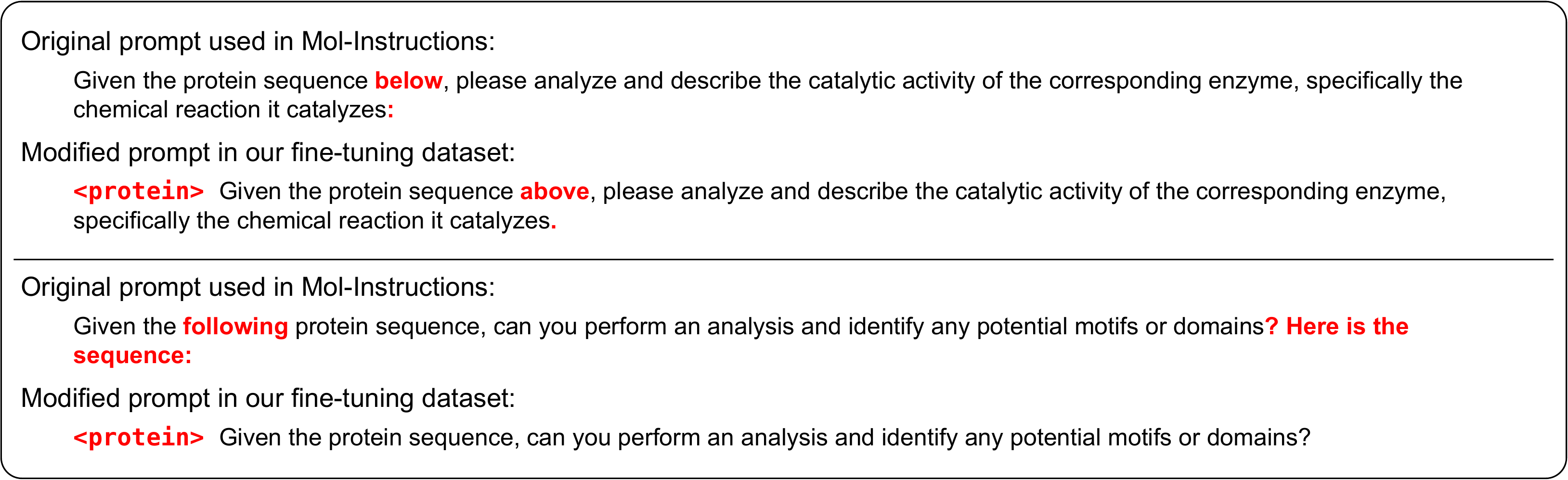}
    \caption{Examples of modifications to the original prompts in Mol-Instructions. \textcolor{red}{Red expressions} are highlighted for modifications.}
    \label{fig:molinst-template}
\end{figure*}

For each task in Mol-Instructions, we make simple modifications to the original prompts to ensure that they are suitable for our use cases and maintain coherence. First, we remove the appended text-format FASTA sequences. Additionally, we modify some expressions in the original prompts. Some modification cases are listed in Fig.~\ref{fig:molinst-template}.

\section{Experimental Setups}
\label{sec:exp-setup}
For protein understanding tasks, we follow Mol-Instructions to split the dataset into an 8:1:1 ratio for training/validation/test, where the training and validation sets are used for the supervised fine-tuning stage, and the test set is used for assessing model performance. For protein property prediction tasks, we follow the PEER benchmark to split the dataset for each task. The details of dataset splits are listed in Tab.~\ref{tab:dataset}. The results are averaged over three runs with different random seeds. Specifically, we follow the PEER benchmark to report the mean and standard deviation of three runs' results.

\input{tables/tab-dataset}

We conduct both the projection tuning stage and supervised fine-tuning stage on 80GB H800 GPUs. The experiments to evaluate the inference latency, reported in Tab.~\ref{tab:ablation-fusion-more}, are conducted on 24GB RTX 3090 GPUs. The hyperparameters are listed in Tab.~\ref{tab:hyperparam}.

\input{tables/tab-hyperparam}

\section{Evaluation Implementation Details}
\label{sec:eval-impl}

\begin{figure*}[!htb]
    \centering
    \includegraphics[width=\linewidth]{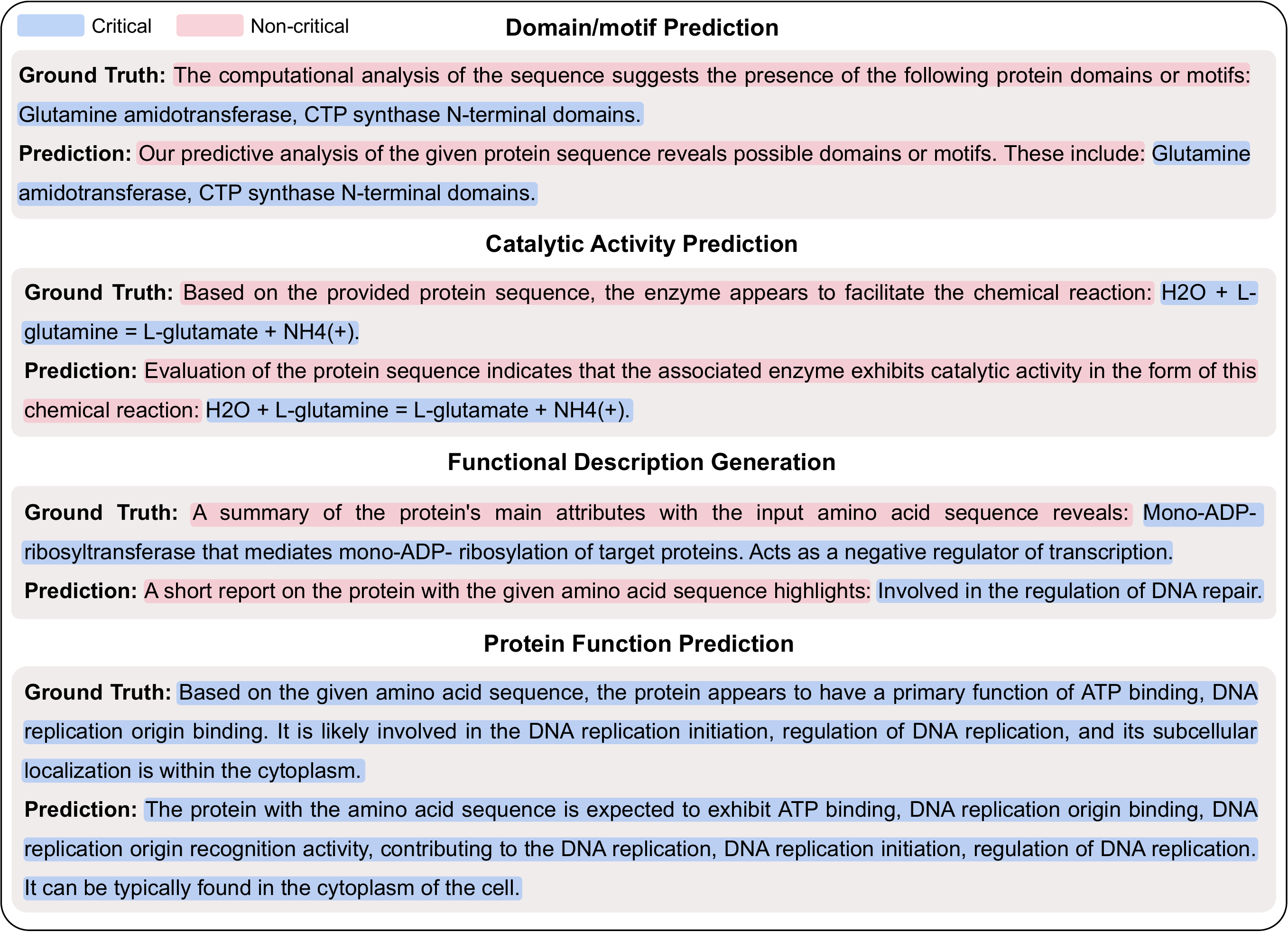}
    \caption{Examples of computing ROUGE-L score on protein understanding tasks.}
    \label{fig:exp-rouge}
\end{figure*}

For a fair comparison, we follow Mol-Instructions \citep{fang2023mol} to compute the ROUGE-L \citep{lin2004rouge} score. Specifically, we take the complete references and predicted answers as inputs. However, both the references and predictions contain some non-protein-related parts, which are non-critical. In Fig.~\ref{fig:exp-rouge}, we show that the critical parts are related with the domains, catalytic activities, and functions. To illustrate how the critical parts affect the ROUGE-L score, we exclude the non-critical parts, retaining only the domains/motifs, chemical reactions, and functional descriptions in both the reference and prediction. Note that the modifications are applied to three sub-tasks, with the exception of the protein function prediction task, where every part of the generation is critical to the protein functions.

The re-evaluation results are displayed in Tab.~\ref{tab:critical-rouge}, demonstrating that by excluding the non-critical parts from both the references and predictions, \mymodel still outperforms the previous state-of-the-art model by 5\%. Compared to the 6\% improvement discussed in Sec.~\ref{sec:molinst}, the impact of non-critical parts is relatively minor.

\input{tables/tab-critical-rouge}

\section{More Evaluations}
\label{sec:more-eval}

We conduct more evaluations on protein property prediction tasks for each ablation study introduced in Sec.~\ref{sec:ablation}. 

\paragraph{Effect of the projection tuning stage}
As shown in Tab.~\ref{tab:ablation-pt}, \mymodel with supervised fine-tuning fails to maintain performance after the projection tuning stage, while \mymodel in zero-shot settings remains competitive with the baselines fine-tuned on PMol (see Tab.~\ref{tab:molinst}). This indicates that bridging the gap between protein structures predicted by AlphaFold-2 and ESMFold during the supervised fine-tuning stage is more challenging than inference.

\input{tables/tab-ablation-pt}

\paragraph{Effect of structure and sequence representations}
As shown in Tab.~\ref{tab:ablation-features}, \mymodel and \mygearmodel enhances the performance on protein understanding tasks by incorporating both structure and sequence representations, specifically on protein function prediction and catalytic activity prediction.

\input{tables/tab-ablation-features}

\paragraph{Effect of the fusion method}
As shown in Tab.~\ref{tab:ablation-fusion}, \mymodel with fused representations outperforms the one without, particularly on subcellular localization prediction task, while \mygearmodel achieves better performance when the structure and sequence representations are not fused. This demonstrates that the fusion method is more effective when ProteinMPNN is used as the structure encoder.

\input{tables/tab-ablation-fusion}

\begin{figure}[!htb]
    \centering
    \includegraphics[width=\linewidth]{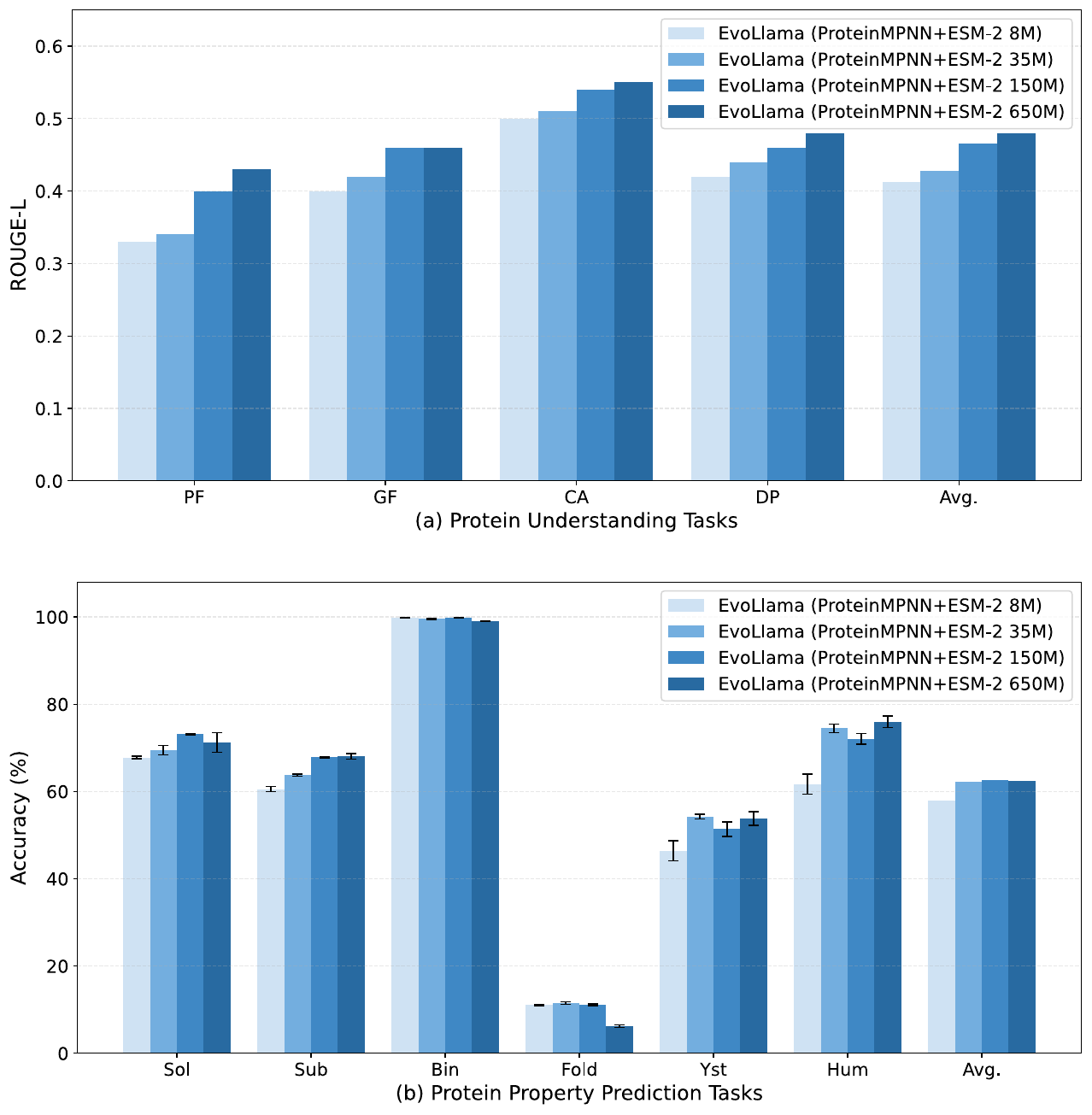}
    \caption{Effect of protein sequence encoder sizes. The experiments are conducted on (a) protein understanding tasks and (b) protein property prediction tasks.}
    \label{fig:ablation-esm}
\end{figure}

\paragraph{Effect of protein sequence encoder sizes}
The ESM-2 650M protein sequence encoder in \mymodel is substituted with encoders of different sizes to demonstrate that the scaling law observed by \citet{lin2022language} extends to our multimodal framework. The experimental results in Fig.~\ref{fig:ablation-esm}(a) indicate that performance on protein understanding tasks improves as the size of the protein sequence encoder increases. Furthermore, \mymodel employing ESM-2 with only 8M parameters outperforms Llama-2-7B-Chat fine-tuned on PMol by an average of 27\%. This result highlights the effectiveness of our approach, as the protein sequence encoder effectively captures evolutionary knowledge from amino acid sequences, substantially enhancing the LLM's understanding of proteins. We also conduct experiments on protein property prediction tasks, as illustrated in Fig.~\ref{fig:ablation-esm}(b). The results show positive accuracy scaling across most tasks, with the exception of fold classification. A possible explanation is that the limited amount of training data makes it challenging for our multimodal architecture to effectively learn the distinctions among the 1,195 fold levels.

\section{Case Study}
\label{sec:case-study}

\begin{figure*}[!htb]
    \centering
    \includegraphics[width=\textwidth]{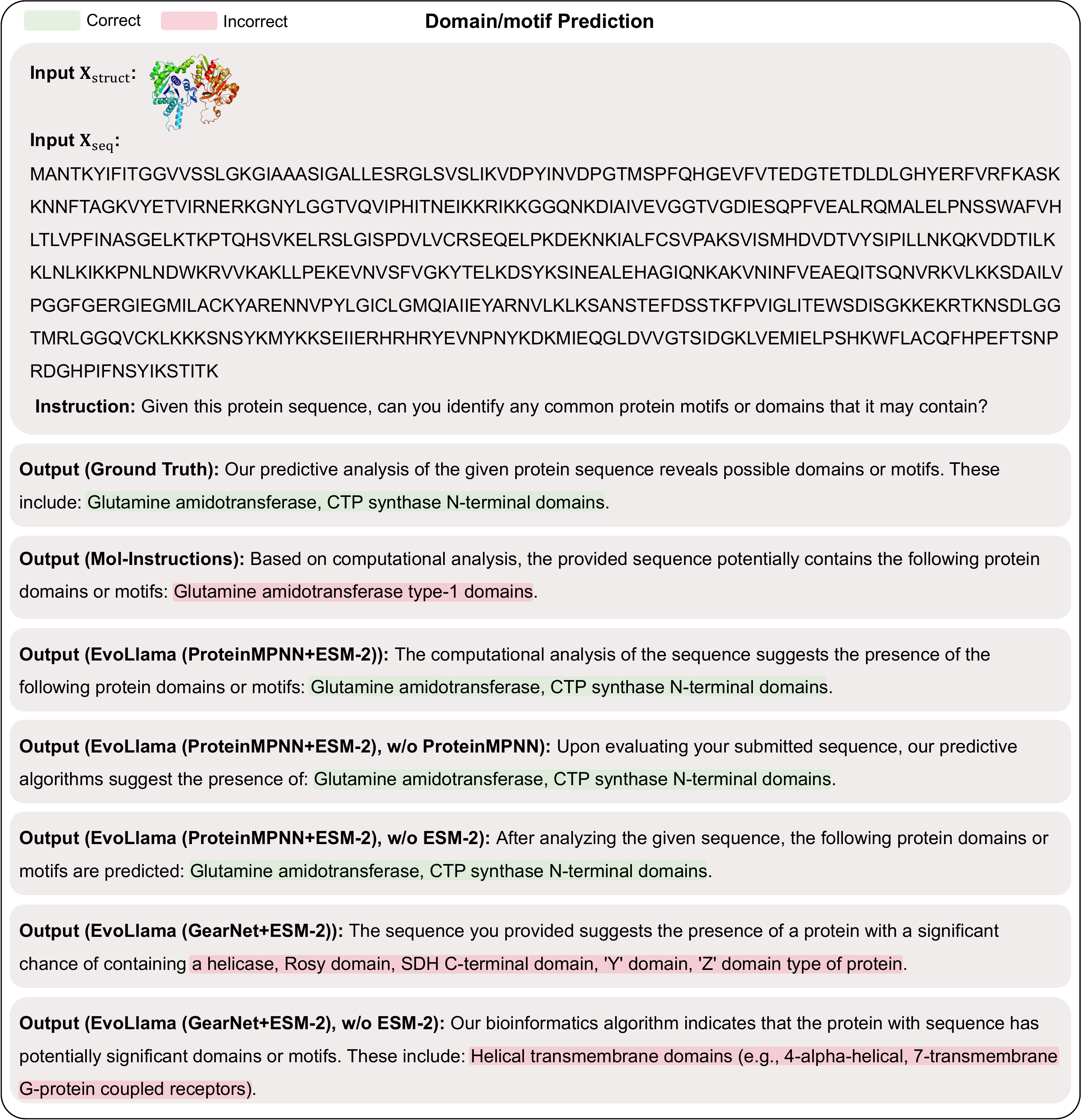}
    \caption{Case study of performance on protein understanding tasks (Domain/motif prediction).}
    \label{fig:case-study-domain}
\end{figure*}

As shown in Fig.~\ref{fig:case-study-domain}, we compare the outputs of \mymodel, \mygearmodel and Mol-Instructions on domain/motif prediction task. Only \mymodel correctly predicts all possible domains or motifs of the given protein, regardless of whether structure or sequence representations are incorporated. Compared to \mymodel, Mol-Instructions fails to predict all the domains or motifs while \mygearmodel generates incorrect ones, indicating that the structure representations extracted by GearNet fail to capture domain- or motif-related information.

\begin{figure*}[!htb]
    \centering
    \includegraphics[width=\textwidth]{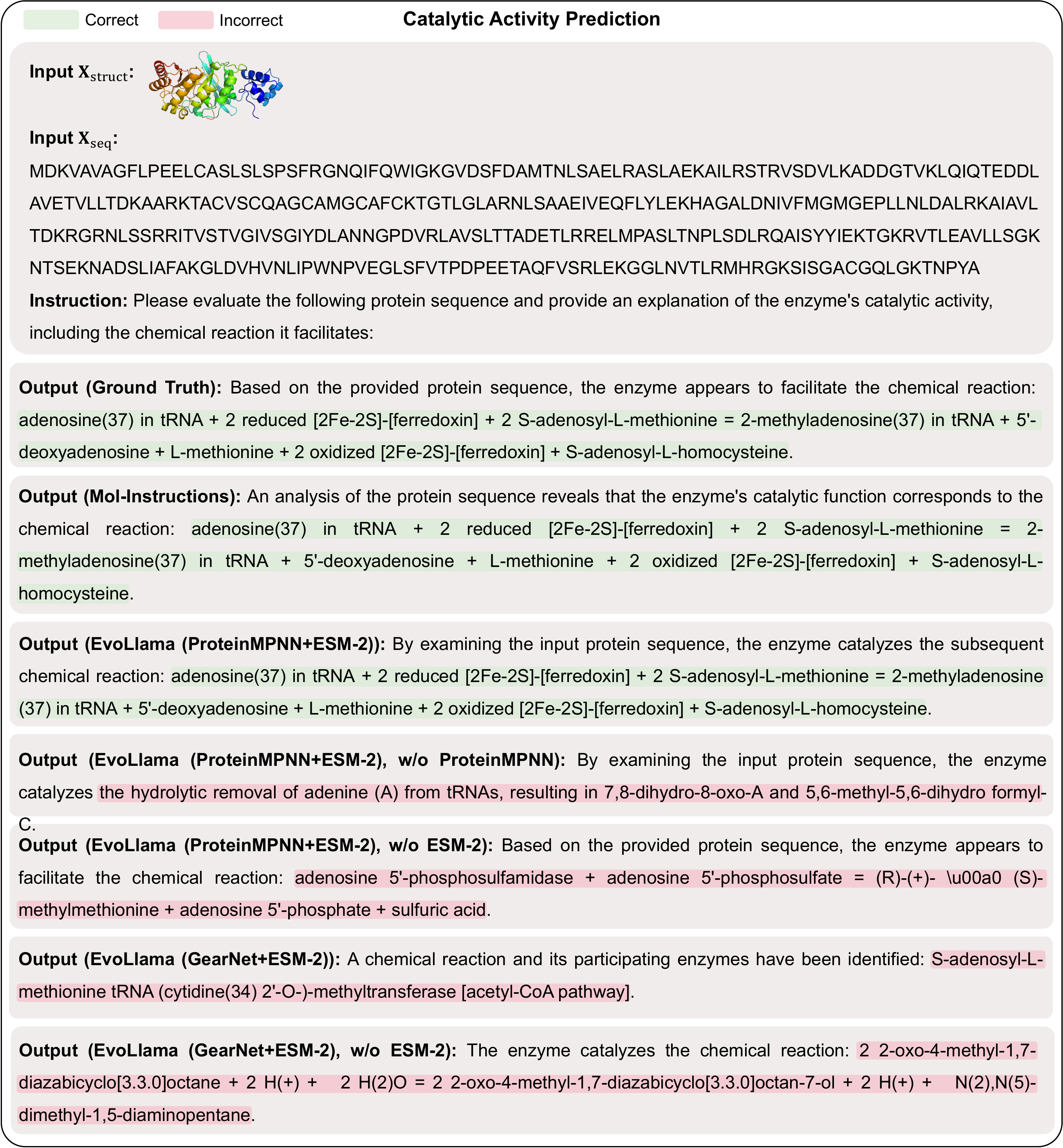}
    \caption{Case study of performance on protein understanding tasks (Catalytic activity prediction).}
    \label{fig:case-study-catalytic}
\end{figure*}

As shown in Fig.~\ref{fig:case-study-catalytic}, we compare the outputs of these models on catalytic activity prediction task. Both \mymodel, with fused structure and sequence representations, and Mol-Instructions generate the accurate and complete chemical reaction. Furthermore, models with only structure or sequence representations fail to produce the correct chemical reaction, demonstrating the significance of fused representations.

%% file: tables/tab-dataset.tex
\begin{table*}
    \centering
    \begin{tabular}{@{}cccccc@{}}
        \toprule
        Tasks & Sub-tasks & Data Source & \#Training & \#Validation & \#Test \\ \midrule
        \multirow{6}{*}{\begin{tabular}[c]{@{}c@{}}Protein\\ Understanding\\ Tasks\end{tabular}} & Solubility & \multirow{6}{*}{\begin{tabular}[c]{@{}c@{}}PEER\\ Benchmark\end{tabular}} & 62,478 & 6,942 & 1,999 \\
         & Subcellular Localization &  & 8,420 & 2,811 & 2,773 \\
         & Binary Localization &  & 5,184 & 1,749 & 1,749 \\
         & Fold Classification &  & 12,312 & 736 & 718 \\
         & Yeast PPI &  & 9,890 & 190 & 788 \\
         & Human PPI &  & 71,338 & 630 & 474 \\ \midrule
        \multirow{4}{*}{\begin{tabular}[c]{@{}c@{}}Protein Property\\ Prediction\\ Tasks\end{tabular}} & Protein Function & \multirow{4}{*}{\begin{tabular}[c]{@{}c@{}}Mol-Instructions\\ (PMol)\end{tabular}} & \multicolumn{2}{c}{110,689} & 3,494 \\
         & Catalytic Activity &  & \multicolumn{2}{c}{51,573} & 1,601 \\
         & Domain/Motif &  & \multicolumn{2}{c}{43,700} & 1,400 \\
         & Functional Description &  & \multicolumn{2}{c}{83,939} & 2,633 \\ \bottomrule
    \end{tabular}
    \caption{Details of dataset splits for supervised fine-tuning data.}
    \label{tab:dataset}
\end{table*}

%% file: tables/tab-hyperparam.tex
\begin{table*}[!htb]
    \centering
    \begin{tabular}{cccccc}
      \toprule
       Stages & lr & Scheduler & Optimizer & \#Batch Size & \#Epochs/\#Steps \\ \midrule
        Projection Tuning & \(2\times10^{-4}\) & cosine & AdamW & 64 & 2 epochs \\ 
        Supervised Fine-tuning & \(2\times10^{-5}\) & cosine & AdamW & 32 & 25,000 steps \\ \bottomrule
    \end{tabular}%
    \caption{Hyperparameters for the projection tuning stage and supervised fine-tuning stage.}
    \label{tab:hyperparam}
\end{table*}

%% file: tables/tab-critical-rouge.tex
\begin{table}[!htb]
    \centering
    \resizebox{\linewidth}{!}{%
    \begin{tabular}{lccccc}
    \toprule
        Models & PF & GF & CA & DP & Avg. \\ \midrule
         & \multicolumn{5}{c}{ROUGE-L} \\
        \cmidrule{2-6}
        Llama-2-7B-Chat & 0.42 & 0.44 & 0.52 & 0.46 & 0.4600 \\
        \mymodel & 0.48 & 0.50 & 0.60 & 0.50 & \textbf{0.5200} \\ \midrule
         & \multicolumn{5}{c}{ROUGE-L (w/o non-critical parts)} \\
        \cmidrule{2-6}
        Llama-2-7B-Chat & 0.42 & 0.46 & 0.56 & 0.57 & 0.5025 \\
        \mymodel & 0.48 & 0.42 & 0.60 & 0.71 & \textbf{0.5525} \\ \bottomrule
    \end{tabular}%
    }
    \caption{Re-evaluation on protein understanding tasks. The ROUGE-L score is computed excluding the non-critical parts from both the references and predictions.}
    \label{tab:critical-rouge}
\end{table}

%% file: tables/tab-ablation-pt.tex
\begin{table*}[!htb]
    \centering
    \resizebox{\textwidth}{!}{%
    \begin{tabular}{lcccccccc}
    \toprule
        Models & Sol & Sub & Bin & Fold & Yst & Hum & Avg. \\ \midrule
        \mymodel & 71.19\(\pm\)2.27 & 68.05\(\pm\)0.65 & 99.10\(\pm\)0.05 & 6.18\(\pm\)0.29 & 53.81\(\pm\)1.55 & 75.95\(\pm\)1.30 & \textbf{62.38} \\
        \begin{tabular}[c]{@{}l@{}}\mymodel \\ ~ ~ w/ PT \end{tabular} & 62.28\(\pm\)0.25 & 50.73\(\pm\)0.11 & 92.37\(\pm\)0.35 & 2.14\(\pm\)0.92 & 50.30\(\pm\)0.22 & 65.54\(\pm\)0.99 & 53.89 \\ \bottomrule
    \end{tabular}%
    }
    \caption{More evaluations on the effect of the projection tuning stage. The experiments are conducted on protein property prediction tasks.}
    \label{tab:ablation-pt}
\end{table*}

%% file: tables/tab-ablation-features.tex
\begin{table*}[!htb]
    \centering
    \resizebox{\textwidth}{!}{%
    \begin{tabular}{lccccccc}
    \toprule
        Models & Sol & Sub & Bin & Fold & Yst & Hum & Avg. \\ \midrule
        \mymodel & 71.19\(\pm\)2.27 & 68.05\(\pm\)0.65 & 99.10\(\pm\)0.05 & 6.18\(\pm\)0.29 & 53.81\(\pm\)1.55 & 75.95\(\pm\)1.30 & \textbf{62.38} \\
        \begin{tabular}[c]{@{}l@{}}\mymodel \\ ~ ~ w/o ProteinMPNN \end{tabular} & 70.91\(\pm\)0.10 & 68.63\(\pm\)0.21 & 99.73\(\pm\)0.03 & 7.94\(\pm\)0.11 & 54.48\(\pm\)0.60 & 65.12\(\pm\)2.76 & 61.14 \\
        \begin{tabular}[c]{@{}l@{}}\mymodel \\ ~ ~ w/o ESM-2 \end{tabular} & 63.06\(\pm\)0.24 & 40.56\(\pm\)0.56 & 99.41\(\pm\)0.03 & 6.78\(\pm\)0.36 & 53.34\(\pm\)0.61 & 52.74\(\pm\)0.60 & 52.65 \\ \midrule
        \mygearmodel & 67.20\(\pm\)0.40 & 37.96\(\pm\)0.33 & 91.96\(\pm\)0.06 & 3.67\(\pm\)0.29 & 52.16\(\pm\)0.63 & 60.41\(\pm\)0.10 & \textbf{52.23} \\
        \begin{tabular}[c]{@{}l@{}}\mygearmodel \\ ~ ~ w/o ESM-2 \end{tabular} & 57.63\(\pm\)0.07 & 12.15\(\pm\)0.18 & 91.96\(\pm\)0.10 & 3.90\(\pm\)0.20 & 49.58\(\pm\)0.74 & 48.59\(\pm\)0.44 & 43.97 \\ \bottomrule
    \end{tabular}%
    }
    \caption{More evaluations on the effect of structure and sequence representations. The experiments are conducted on protein property prediction tasks.}
    \label{tab:ablation-features}
\end{table*}

%% file: tables/tab-ablation-fusion.tex
\begin{table*}[!htb]
    \centering
    \resizebox{\textwidth}{!}{%
    \begin{tabular}{lccccccc}
    \toprule
        Models & Sol & Sub & Bin & Fold & Yst & Hum & Avg. \\ \midrule
        \mymodel & 71.19\(\pm\)2.27 & 68.05\(\pm\)0.65 & 99.10\(\pm\)0.05 & 6.18\(\pm\)0.29 & 53.81\(\pm\)1.55 & 75.95\(\pm\)1.30 & \textbf{62.38} \\
        \begin{tabular}[c]{@{}l@{}}\mymodel \\ ~ ~ w/o fused representations \end{tabular} & 69.53\(\pm\)0.87 & 59.29\(\pm\)2.35 & 99.16\(\pm\)0.50 & 10.91\(\pm\)0.37 & 56.26\(\pm\)1.09 & 74.40\(\pm\)0.70 & 61.59 \\ \midrule
        \mygearmodel & 67.20\(\pm\)0.40 & 37.96\(\pm\)0.33 & 91.96\(\pm\)0.06 & 3.67\(\pm\)0.29 & 52.16\(\pm\)0.63 & 60.41\(\pm\)0.10 & 52.23 \\
        \begin{tabular}[c]{@{}l@{}}\mygearmodel \\ ~ ~ w/o fused representations \end{tabular} & 64.33\(\pm\)0.00 & 47.66\(\pm\)0.87 & 91.90\(\pm\)0.27 & 6.59\(\pm\)0.23 & 56.56\(\pm\)1.72 & 70.04\(\pm\)0.86 & \textbf{56.18} \\ \bottomrule
    \end{tabular}%
    }
    \caption{More evaluations on the effect of the fusion method. The experiments are conducted on protein property prediction tasks.}
    \label{tab:ablation-fusion}
\end{table*}